# Real-Time Robust Tracking of Moving Robots with Multiple RGBD Consumer Cameras

Abdenour Amamra and Nabil Aouf
Department of Informatics and Systems Engineering, Cranfield University, UK
{a.amamra, n.aouf}@cranfield.ac.uk

*Abstract*—This paper presents a new approach to track moving vehicles with multi-view consumer RGBD cameras. Our solution is able to accurately issue position and orientation data of rigid bodies in occluded indoor environments. Its architecture is based on a pipeline taking as input the raw image data issued by the sensors, upon which we localise the robots in their space. Before being available, the 3D position of the vehicle is subject to a sensor-wise filtering scheme. Afterwards, a data fusion algorithm is applied to merge the individual outputs of each camera in a single trajectory characterising the real world positions and orientations of the moving vehicles. To reach real-time performance, we benefited from the high parallel computational capability of the graphic processors. Our system works at a frequency of twenty five frames per second for five cameras. Test results show the accuracy we reached and the effectiveness of our solution.

## I. Introduction

In recent years, real-time tracking of moving objects in RGB video streams becomes one of the most targeted research areas in robotics and computer vision. Surveillance, sports reporting, video annotation, and traffic management systems are few of the domains which benefited from advances in this field [1]. Although the higher performance achieved until now, RGB data is limited in giving a complete knowledge about the real world to just colour and intensity. Information about the 3D structure can be recovered by many hardware means though. New off the shelf RGBD sensors like Microsoft Kinect opened wide insights on a better perception of the surrounding space at a compelling frequency of capture [2]. These cameras have the ability to deliver both the 3D map of the scene along with the corresponding colour information.

In the present work, we focus on the use of multiple RGBD sensors to accurately localising moving robots. The result is used to feed augmented reality and robotic systems with real-time position data. Single camera tracking is hampered by the camera's limited field of view. However, multiview setup is better to overcome occlusions among different viewpoints and to benefit from the joint action of all the sensors for a more accurate tracking. Nevertheless, larger amounts of 3D data yield extra load of computations that may inversely affect the response time during the scenario. Hence the real-time nature of the system requires us to find a compromise between performance and response time throughout the whole processing line. Graphic processing units (GPU) are a very powerful tool when the algorithm could be designed in a parallel architecture [3]. In our specific case, the huge amount of 3D data simultaneously output by the multiview setup is subject to a filtering and a fusion algorithm. The processing pipeline starts with RGBD data acquisition from all the sensors at the same time. These data are then sent through Kalman filtering stage to improve their precision [4]. 3D positions of the targets are then computed and therefor sent to robust Kalman/$H_\infty$ filtering stage to correct them. Based on sensor-wise estimates, a data fusion algorithm is afterwards applied to combine all position data in one decent output. This algorithm uses a set of weighting coefficients computed upon the accuracy of estimates and the uncertainty in the measurements of each camera. Our main contributions are as follows:

- We improved the raw measurements of RGBD sensor using a mathematical correction model appropriate to each sensor.

- We dealt with the uncertainties in the model describing the motion of the vehicle using robust Kalman/$H_\infty$ filter.

- We applied covariance intersection algorithm with adaptive weighting coefficients computed from the specific properties of our tracking setup.

The paper is structured as follows: in **related works** section we discuss the state of the art of image based tracking applications and their use in robotics and augmented reality. We explain the **architecture of our system** in the following section. We then give the details about the first two modules of the system in **capture and markers extraction** section. In **Robust filtering** section, we explain the modelling of uncertainties and how we were able to accurately track the objects without an exact knowledge about their motion model. The last step of our pipeline is covariance intersection technique in which all the filtered outputs are fused to form a unique pose and position associated to each vehicle present in the scene. To validate our finding, we present the results we got from real experiments in **results and discussions** section. Finally, we conclude by showing the pros and the cons of our method along with possible improvements in future works.

## II. RELATED WORKS

Tracking problem is generally divided into three stages: motion detection, object segmentation, and object tracking [5]. Single-camera tracking methods, mostly suffer from object-object or object-obstacle occlusions, the latter lead to failure as the tracked entities become wrongly associated [6]. Zhao et al. [7] presented a method for tracking multiple people with a single camera. They used 3D shape models of people that were projected back in image space to perform the segmentation and to resolve occlusions. Each human hypothesis was then tracked in 3D with Kalman filter using the object's appearance constrained by its shape. Okuma et al. [8] propose a combination of Adaboost for object detection and particle filters for multiple objects tracking. The combination of the two approaches leads to fewer failures than either one on its own, as well as addressing both detection and consistent track formation in the same framework. Leibe et al. [9] present a pedestrian detection algorithm for crowded scenes. Their method iteratively aggregates local and global patterns for better segmentation. These and other similar algorithms are challenged by occluding and partially occluding objects and appearance changes. On the other hand, multiview object tracking has recently benefited from a large interest against single camera alone [6]. This advantage is largely driven by the wider coverage of the scene which is an asset in handling occlusions. KidsRoom system developed at MIT Media laboratory [10] used a real-time tracking algorithm based on contextual information. The tracking algorithm uses the overhead cameras view of the space, which minimizes the possibility of one object occluding another. The system could track and analyse the actions and interactions of people and objects. Lighting is assumed to remain constant during runtime. Background subtraction is used to segment objects from the background [11], and the foreground pixels are clustered into 2D blobs. The algorithm then maps each person known to be in the room with a blob in the incoming image frame. Pfinder (Person-finder) is another real-time system for the tracking and interpretation of human motion [12]. Motion detection is performed using background subtraction, where the statistics of background pixels are recursively updated using a simple adaptive filter. The human body is modelled as a connected set of blobs using a combination of spatial and colour cues. Pfinder has been applied for a variety of applications including: video games, distributed virtual reality, providing interfaces to information spaces, and recognising sign language. In our system, we only track robots as a first stage. We further apply background subtraction to extract the position of the markers stuck on the vehicle. However, the computation of the actual centre of mass is based on the extraction of contours for every marker and the computation of its corresponding zeroth and first moments [13]. This stage is applied on the individual outputs of each camera separately.

From a filtering point of view, the tracking problem is considered as sequential recursive estimation. The estimation combines the knowledge about the previous estimate and the current measurement using a state transition model. In image space, the measurement comes from the relative 3D position of the vehicle in the space where it is evolving. Each frame is processed within a time step. Noise statistics are issued from noise processes affecting both the measured and estimated positions. The state-space formalism, where the current tracked object properties are described in an unknown state vector updated by noisy measurements is very well adapted to model tracking. The sequential estimation has an analytical solution under very restrictive hypothesis. Kalman filter (KF) is an optimal solution for the class of linear Gaussian estimation problems [14], On the other hand, for nonlinear systems, a number of Bayesian techniques have been proposed to perform the optimization. When Gaussian distribution is assumed, commonly used approaches include the extended Kalman filter (EKF) [15] and the Unscented Kalman filter (UKF) [16]. EKF computes the Jacobian matrices and derives a linear approximation to the nonlinear function. While the UKF avoids computing the Jacobian matrices and propagate the distribution by sampling. Furthermore, particle filter also avoids the computation of the Jacobian matrices, it is a numerical method that allows finding an approximate solution to the sequential estimation[17]. All the filters above are very sensitive to error in the system's model. In other words, if the system is imprecisely modelled, which is very common with real scenarios [18], the filter behaves wrongly. To overcome the lack of knowledge about the system, we consider in the present paper using robust Kalman/$H_\infty$ (RF) filter which has the ability to cope with uncertainties in the system's model. Up to our knowledge, none investigated the adaptation of RF for accurate tracking before. Furthermore, Covariance intersection (CI) [19] is applied to combine the estimates computed upon the raw output of each camera in a way that minimises the error in the global estimate [20].

RGBD sensors were recently very famous among professional and research works alike. They do not only deliver colour but also depth information available at a frame rate of 30Frames/second with Kinect. This information (colour and depth) enables us to benefit from both image and 3D geometry approaches to overcome the traditional problems of many robotic and computer vision applications, such as human pose estimation [21], robot navigation [22], SLAM [23], and object tracking [24], 3D scanning [25]. However, the data resulting from these RGBD sensors are more important in size than their colour counterpart. Thus the need for GPUs to reach real-time performance. Many examples in the literature achieved higher performance when the bottlenecks in processing were correctly implemented on the GPU. Kinect fusion [25] and the work of Tong et al. [26] are the best examples. The innovative point in our work compared to its ancestors is that we use RGBD data delivered by a multiview setup to track moving vehicles. We employed robust filter to overcome the uncertainties about motion model of the robot moving in front of the cameras. As well as covariance intersection algorithm with adequate weighting coefficients to optimally merge the estimates resulting from sensor-wise RF. All the parallelisable parts of our solution are designed to run on the GPU. Hence the high frame rate we reached given the large amount of image data streamed in real-time. This work opens new perspectives on the possibility to use cheap sensors for accurate tracking and the applicability of RF on other problem where the system is not precisely characterised.

TABLE 1
SYMBOLS AND THEIR CORRESPONDING MEANING

| Symbol | Quantity/Unit/Range/Size | Definition |
|---|---|---|
| $N$ | 5 | Number of cameras covering the whole scene |
| $f_x, f_y$ | pixel | Focal lengths in pixel unit toward *X,Y* axes respectively |
| $c_x, c_y$ | pixel | Centre of the imager |
| **R** | $3 \times 3$ | Rotation transform from colour camera (RGB) frame to infrared (IR) frame |
| **T** | $3 \times 1$ | Translation transform describing the shift between RGB frame centre and IR frame |
| $u, v$ | pixel | Coordinates of a given pixel in the IR frame |
| $z(u,v)$ | 0.8 – 4.5 (meter) | Depth measure which represents the distant that separates the sensor from the scene |
| $x, y, z$ | meter | World coordinates of the robot according to the three axes of the world frame |
| $\dot{x}, \dot{y}, \dot{z}$ | meter/second | Velocity of a 3D point according to the three axes of the world frame |
| $\ddot{x}, \ddot{y}, \ddot{z}$ | meter/second$^2$ | Acceleration of a 3D point according to the three axes of the world frame |
| $Im\_Size$ | $640 \times 480$ | Size of images in every RGB and IR frame |
| $Rgbd\_K_i$ | $640 \times 480$ | Raw RGBD (RGB + Depth) data output from the *i*-th *(0≤ i ≤4)* Kinect sensor |
| $KF\_Rgbd\_K_i$ | $640 \times 480$ | Pre-processed and aligned frames (filtered with Kalman filter) |
| $Pos_i$ | $3 + 3$ | Position and orientation of the robot issued from the *i*-th *(0≤ i ≤4)* Kinect sensor |
| $RF\_Rob_i$ | $3 + 3$ | Position and orientation of the robot issued after applying robust Kalman/$H_\infty$ filter |
| $Cov_i$ | $3 \times 3$ | Covariance matrix associated to $RF\_Rob_i$ computed by robust Kalman/$H_\infty$ filter |
| $\overrightarrow{Ox}, \overrightarrow{Oy}, \overrightarrow{Oz}$ | $1 \times 3$ | The orientation of the robot computed from the positions of the markers |

## III. SYSTEM OVERVIEW

### A. Kinect camera

Kinect sensor[1] is an RGBD camera which has the ability to capture the depth map of the scene, and its RGB colour image at a frame rate of 30Hz and a resolution of $640 \times 480$ pixels. The camera holds an IR-projector that projects a set of IR patterns on the scene. An IR-camera captures the light reflected by the projected patterns and an RGB imager senses the colour of objects present in front of the camera [27]. The sensor infers the depth of the scene based on the computation of disparity with a triangulation process. After a calibration procedure, both RGB image and IR depth data can be fused to form a coloured 3D point cloud of about 300,000 points in every frame.

Although Kinect comes with factory embedded calibration parameters ($f_x, f_y, c_x, c_y$ intrinsic parameters for both RGB and IR cameras and the extrinsic parameters [**R**, **T**] of the IR-RGB stereo setup) the actual accuracy of the capture may range from less than one millimetre to many centimetres. This clear difference depends on the state of the sensor, the target application and the nature of the scene [28]. To decently track moving objects with Kinect, the sensor should be accurately recalibrated. Native parameters are more generic and remain the same for all the Kinects in the market. However, the frequency of use and the external factors which widely differ from one application to another, can easily affect the precision of measurements [24]. Hence the performance of the whole tracking module will be affected.

### B. Hadrware and software configuration

Our real-time tracking setup is composed of $N = 5$ Kinect cameras covering a volume of $4 \times 4 \times 3$ metres, all the sensors are connected to the same workstation Figure 1. To respond to this relatively high frame rate, a powerful computing hardware is necessary. Alternatively, we can divide the load as much as possible on many low end computing units; this involves an extra burden to control the flow of data between all the different units.

In our system, we benefited from the power of the graphic processing unit (GPU) to process the considerable amount of image data issued by the five cameras in real-time. The hardware configuration encloses an INTEL i7 3930K CPU holding six physical cores (two logical cores per physical) running at 3.20 GHz, 16.0 GB of RAM along with an NVIDIA GeForce 2GB GTX 680 GPU. In our case of study, we track a ground robot (Pioneer P3-DX[2]). However, our tracking system could be used to issue the 3D trajectory of any kind of ground or aerial vehicles evolving in indoor environment. During the experiment, the robot evolves freely in the space covered by the sensors according to an embedded obstacle avoidance algorithm that runs simultaneously with the capture. For this reason (the vehicle is freely evolving), we used a more general motion model which adapts to every moving entity within the space covered by the set of cameras.

From a software point of view, we need to access in real time the output of all the Kinects simultaneously. Thus, we used Kinect SDK 1.7.0[3] which comes with the driver of the sensor provided by the manufacturer. To control and program the GPU we used CUDA[4]. CUDA as a GPU/CPU programming language allows us to build heterogeneous applications that will be fragmented into pieces to run on the CPU and some kernels to be launched on the GPU [29].

---

[1] Microsoft Kinect : http://www.xbox.com/en-GB/Kinect, 2012
[2] http://www.mobilerobots.com/researchrobots/pioneerp3dx.aspx, 2013
[3] http://www.microsoft.com/en-gb/download/details.aspx?id=36998, 2013
[4] http://www.nvidia.com/object/cuda_home_new.html, 2013

*C. Real-time multi-Kinects tracking architecture*

The system consists of four main modules through which flows every synchronised set of RGBD data issued by all the five sensors Figure 2, Figure 3:

*1) Capture module:*

It is the responsible on feeding the tracker and accordingly the following stages of the platform with 3D point clouds. This part of the application is multithreaded, that is, every thread is responsible on the processing of the data output by a given sensor. Although the architecture of our solution fits into centralised model (all the processing is done at the level of the same workstation) the multithreaded design increases its flexibility and its scalability to support any number of sensors. More importantly, such approach allows a full occupation of both the CPU and the GPU during the whole time of capture. This outcomes a higher throughput and increases the ability to significantly approach the native frame rate of the sensor. Nevertheless, other sensor related limitations should be taken into account when using multiple Kinects simultaneously. The IR beams emanating from the projectors can interfere with each other, and consequently confuse the IR cameras when inferring the disparity (it would be impossible to decide which IR speckle belongs to which sensor). As a result, some holes appear in the 3D data because of the undefined disparity information resulting from the incapacity of computing the actual value [30].

Capture module receives raw RGBD data from the Kinects as input. Each thread associated to a given sensor will act independently by loading the data into the GPU and performing the following computations:

- Filtering the unreliable $z_i$ elements (*i-th* pixel in the depth map where no disparity information is available).
- Correcting the remaining valid depth values using the appropriate correction model of the sensor.
- Computing $x_i, y_i$ for the valid points only using the intrinsic parameters of the IR camera.
- Applying Kalman filtering scheme [31] to optimally filter the noisy coordinates.
- Mapping the colour image onto the depth image using stereo calibration parameters.

The Capture module is totally implemented on the GPU. In other words, all the steps listed above constitute a pipeline which runs in parallel for all the pixels simultaneously. The output is a consistent 3D coloured point cloud.

*2) Markers extraction module:*

To compute the position and orientation of the robot, we fixed three markers on its top Figure 1. The three markers allow us to get the 3D pose of the moving object by estimating its centre of mass and the corresponding orientation based on RGBD data. This becomes possible because we have the colour data registered with the depth map. Consequently, we should just fetch the markers in the colour space, and then the corresponding 3D positions will be straightforward. The makers we used have a distinctive colour which should not be present in the background to facilitate the segmentation. This module takes as input the aligned colour image and follows these steps:

- RGB to HSV (Hue, Saturation and Value) conversion as HSV space is more robust to light intensity change [32].
- Colour thresholding to separate the markers from the background.
- Erode and dilate the binary thresholded image.
- Actual extraction of the markers.

The extraction algorithm is also implemented on the GPU. The output of this module is the position and orientation of the robot within the processed frame.

*3) Robust filtering module:*

The filtering module aims to enhance the quality of position and orientation information issued by the stages above. It acts on the whole trajectory taken by the moving robot over time, and filters it according to a roughly predefined state-transition motion model. However, for generality and to allow the solution to work with any kind of ground or aerial vehicles, we assume that we do not have an exact model of the moving object and we choose a classical Newtonian motion system with approximate parameters.

To overcome the loose fit of this general model to the real system, we applied robust Kalman/$H_\infty$ filtering scheme [33]. This filter deals with the uncertainties in the system and measurement matrices. For the present case of study, we could compensate the lack of knowledge about the nature of system by adapting the filter to a Newtonian model of motion. This module is implemented on the GPU and its output is filtered object position and orientation resulting from every camera alone. The filter is applied on the tracking data at a given time-step (the actual frame which is being processed by the pipeline).

*4) Data fusion module:*

Lastly, after getting each sensor's filtered position and orientation data from the precedent algorithm, we proceed to the combination of all the information to produce a complete and more accurate position data. Prior to this stage, the output of every camera was treated separately by the corresponding thread as if it was the only sensor in the system. However, at this level we use the external calibration parameters of the Kinects with the reference frame to combine the data with covariance intersection technique [34][19]. According to the following steps:

- Transforming each position data for every Kinect to a global reference space.
- Applying covariance intersection algorithm on these data to produce a unique decent position and orientation information.

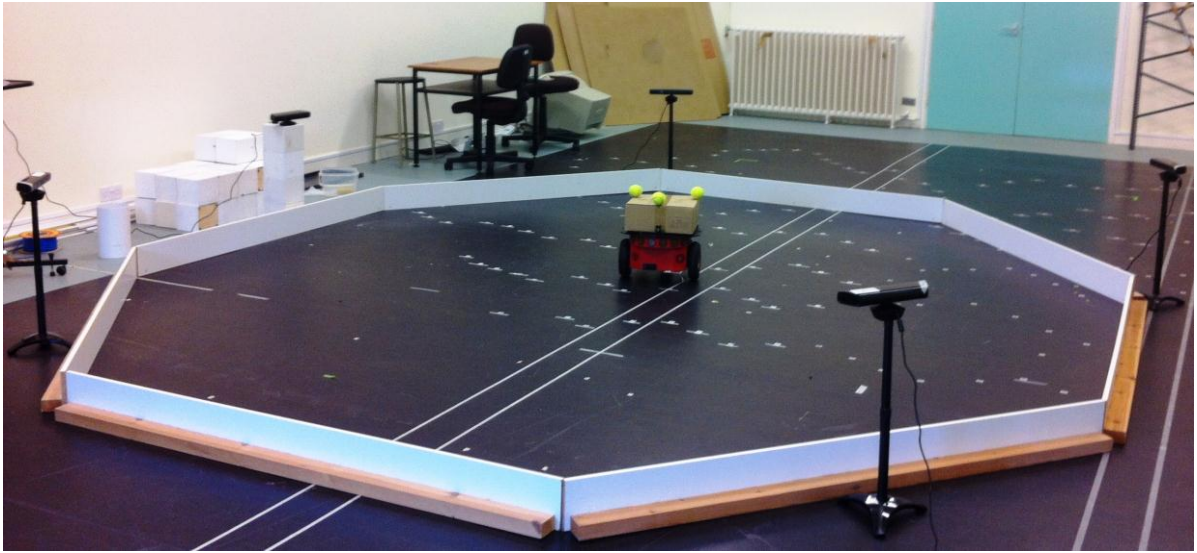

Figure 1. Multi-Kinects real-time tracking system

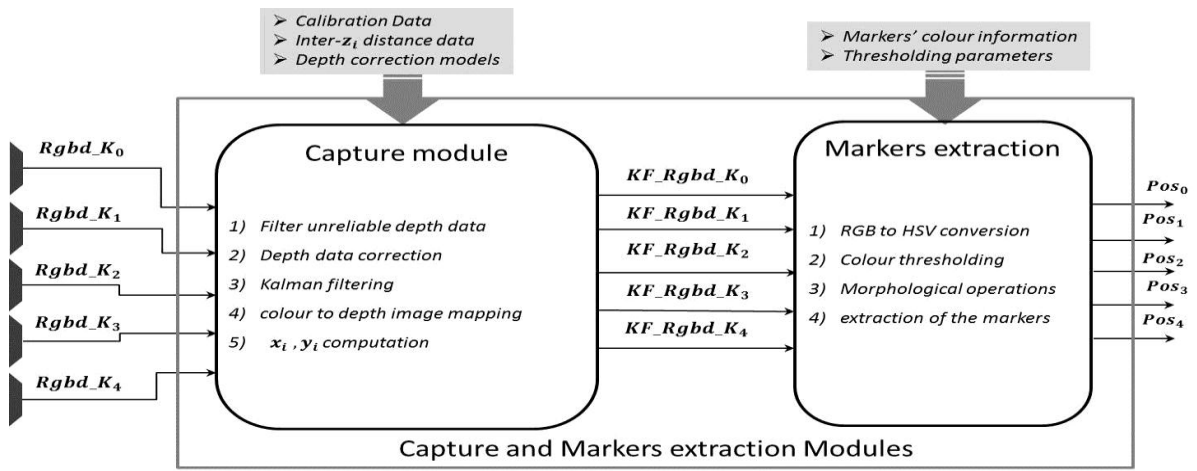

Figure 2. Capture and markers extraction modules

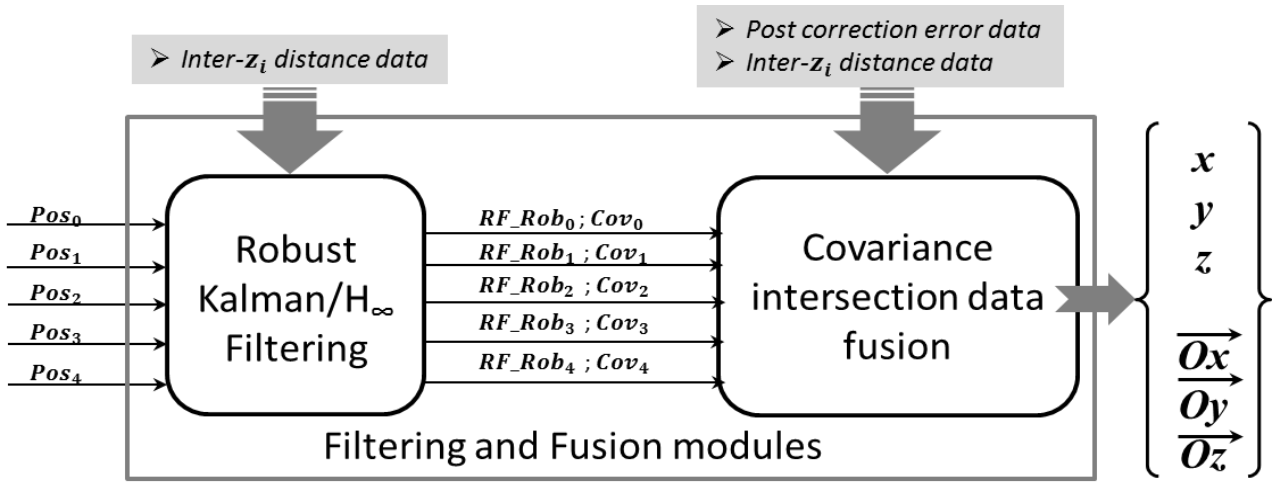

Figure 3. Filtering modules

As the markers can be occluded during the tracking, it is necessary to compensate the missing data. In addition the quality of the measurement is not the same among the five viewpoints. Thus a weighting coefficient is associated to each sensor according to the error in its measurements. These coefficients favour the more accurate estimate based on covariance matrices resulting from the filtering module. This approach allows us to optimally fuse all the outputs of the sensors in a more precise estimate benefitting from the best of every sensor.

## IV. CAPTURE AND MARKERS EXTRACTION

### A. Capture module

*1) Filtering the unreliable $z_i$*

Kinect's driver produces the depth map based on the raw disparity data issued by the device. The IR camera/projector setup forms a stereo pair with a baseline of approximately 7.5 cm. The projector emits a beam of known IR patterns on the scene. These patterns are generated from a set of diffraction gratings, with special care to reduce the effect of zero-order propagation at the centre bright dot [35]. The actual depth is computed by a triangulation process correlating each measured value to a reference disparity stored in the device. In other words, for each pixel in the IR image, a small correlation window is used to compare the local pattern at that pixel with the reference one. The best match gives an offset from the known depth, which is the disparity measure. Kinect performs a further interpolation of the best match to reach sub-pixel accuracy. The camera computes the disparity according to equation (1):

$$d = \frac{1}{8}(doff - kd) \qquad (1)$$

Where $\boldsymbol{d}$ is the normalized disparity, $\boldsymbol{kd}$ is Kinect's disparity, and $\boldsymbol{doff}$ is an offset value particular to the device. The factor $\boldsymbol{1/8}$ appears because the values of $\boldsymbol{kd}$ are in $\boldsymbol{1/8}$ pixel units[5].

After getting the depth map, our capture stage begins with the filtering of unreliable $z_i$ to reduce markers search space in the corresponding RGB image. A depth value is considered unreliable if:

- There is no disparity information at the raw depth pixel.
- It exceeds the interval of useful depth data (0.8m – 4.5m).

*2) $z_i$ correction with sensor's model*

After being used many times, every electronic device suffers from a decrease in accuracy. Kinect as an off the shelf RGBD camera obtains the depth of the scene from the active IR projector/camera setup. The quality of the 3D map issued by the sensor is highly affected by the performance of this IR setup. However, the RGB camera (passive part of the device) is more robust and the colour image remains unaffected for longer time. Along our research, we noticed that larger errors in the depth measure are more likely to appear within the more frequently used sensors. On the other hand, the less used devices do not suffer as much drop in quality of their measurements. This fact means that they produce a more descent and correct 3D data. As a consequence, to ensure a higher precision in our tracking system, and to reduce the negative effect of the worn sensors, we proposed an effective correction approach to enhance the quality of measurements among different Kinects covering the same scene. Up to our knowledge, none has discussed this issue of accuracy in Kinect sensor yet. To rectify the erroneous measure, we used a mathematical model proper to each sensor. The design of such model is based on finding the appropriate function which takes as input the shifted raw output of the device, and accordingly computes the corrected values which hold lower error. Such approach is justified by the fact that regular camera calibration procedure cannot overcome this drop in the quality of disparity measurement. The origin of the problem is not related to the optical configuration of the infrared camera, but to an error factor generated by disparity computation module [36].The latter is an offset that appears in the estimation of the distance between the sensor and the object. Its value becomes more important as the object gets further away. Our idea is based on finding a correction function. Such function should be able to correctly remap the shifted depth value ($\boldsymbol{z_{sh}}$) to its respective world position. Hence, we took the depth measures output by the sensor and their corresponding ground truth ones ($\boldsymbol{z_{cor}}$), the pairs ($\boldsymbol{z_{sh}}, \boldsymbol{z_{cor}}$) are related by a function $\boldsymbol{f(z_{sh}) = z_{sh} - z_{cor}}$ (the shift from the real value). $\boldsymbol{f(z_{sh})}$ issues for each depth value the corresponding error based on precise ground truth reference. $\boldsymbol{f(z_{sh})}$ is computed from sample points taken simultaneously from Kinect and a high precision tracking system[6]. Based on the pairs ($\boldsymbol{z_{sh}}, \boldsymbol{z_{cor}}$) we fit the sparse data with a polynomial that better approaches the shape of the representative curve. The same polynomial will serve as a function which takes as input a raw measurement ($\boldsymbol{z_{sh}}$) generated by Kinect and outputs a corrected estimate ($\boldsymbol{z_{cor}}$) that better approaches the real range separating the object from the camera. Although this helps clearly overcoming the problem of shift in the depth data, the drop in resolution characterising the sensor could not be resolved. In other words, the correction algorithm improves the quality of the device, but it cannot increase the resolution as it is hardware limited. The proposed correction method deals with the decrease in resolution in the same way as the elimination of the offset. The algorithm of the depth map is implemented on the GPU and runs simultaneously with the capture. Consequently, the measurements are corrected without extra computation load on the capture module.

*3) Kalman filtering of the $z_i$*
*a. Kalman filter*

When we point the sensor against a static scene, and we observe the depth map over time (both the sensor and the scene remain steady during the whole capture), we notice fluctuations in almost 90% of all map's elements. These fluctuations are explained by the fact that the depth pixels tend to change their values over time [37].

---

[5]http://wiki.ros.org/kinect_calibration/technical, 2013
[6]http://www.naturalpoint.com/optitrack/, 2013

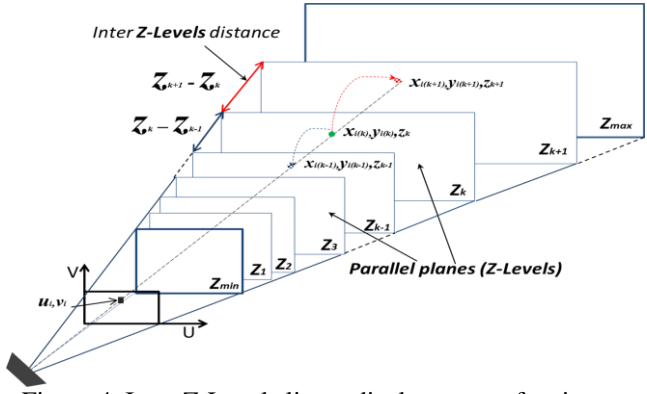

Figure 4. Inter Z-Levels linear displacement of a given point towards the perspective frustum lines

This knowledge about Kinect allows us to benefit from Kalman filter which is an effective tool to produce optimal estimates from a series of noisy measurements [31]. The filter approaches a mean value around which are distributed the noisy raw outputs of the sensor according to their corresponding depth readings. In our system, the filter was implemented on the GPU, it works in real-time (at the same time of the capture) and uses only a small knowledge about the last state (the estimate and its covariance matrix). We successfully adapted Kinect's depth data to Kalman's model in a previous work [37]. The equations are as follows:

*Prediction:*

$$\overline{Z}_k = Z_{k-1} \quad (2)$$

$$\overline{P}_k = P_{k-1} \quad (3)$$

*Correction:*

$$K_k = \overline{P}_k(\overline{P}_k + R_k)^{-1} \quad (4)$$

$$Z_k = \overline{Z}_k + K_k(\tilde{Z}_k - \overline{Z}_k) \quad (5)$$

$$P_k = (1 - K_k)\overline{P}_k \quad (6)$$

For a given pixel in a frame at time-step $k$; $\overline{Z}_k$: is the estimated depth; $\tilde{Z}_k$: measured depth; $\overline{P}_k$: priori estimate error variance and $K_k$: kalman gain.

When the sensor captures a point cloud, the resulting 3D data are automatically distributed on the discrete Z-Levels Figure 5. However, the original points' data come from the continuous real world. Their corresponding images in Kinect's space lie in the parallel planes. The error in measurement is proportional to the gap between Z-Levels Figure 6. Nonetheless, Kalman filter takes these noisy raw data as input, optimises them, and consequently approaches the best their real world counterparts Figure 7.

Figure 8 depicts a real trajectory grabbed by the same device. The blue graph represents the raw trajectory (resulting from Kinect without any filtering). If we observe closely, the deeper are the points (greater $z_i$) the clearer are the parallel Z-Levels (because of the drop in resolution as we get far away from the camera). However, Kalman filter's smoothing effect (Figure 9) appears as it optimally condenses the sparse and discrete $z_i$ around their corresponding real world true positions. Consequently, the error becomes lower.

The filter runs in parallel for all the pixels available in the depth map. This implementation on the GPU was possible because every element is independent from all its neighbours.

*b. $x_i$, $y_i$ computation and stereo mapping*

Kinect has two cameras, one for capturing colour image and the other for capturing IR depth data. Although real-time depth information is delivered by the IR camera, the depth map provides us with how far the objects are from the sensor. Whereas, we do not actually know the depth information of a given pixel in the colour image because the two cameras have different viewpoints. The correct correspondence implies the need for an accurate stereo calibration to get the right [**R, T**] transformation which relates both cameras.

Once the depth data $z_i$ is filtered by Kalman scheme, we proceed to the computation of $x_i$, $y_i$ equations (7), (8) world coordinates. To complete this step, we need the calibration parameters of the IR camera to project the image points to the world frame of the IR camera:

$$x_i = (u_{i\_ir} - c_{x\_ir})\, z_i / f_{x\_ir} \quad (7)$$
$$y_i = (v_{i\_ir} - c_{y\_ir})\, z_i / f_{y\_ir} \quad (8)$$

As shown in (Figure 10), we use the stereo calibration parameters [**R, T**] to map the point $P(x_i, y_i, z_i)$ from the IR coordinate system to the RGB one $P'(x'_i, y'_i, z'_i)$ equation (9).

$$P' = RP + T \quad (9)$$

Afterwards, we re-project $P'$ to the RGB imager using the intrinsic parameters of the colour camera equations (10), (11). This is done to complete the correspondence between the 3D points and their colour information.

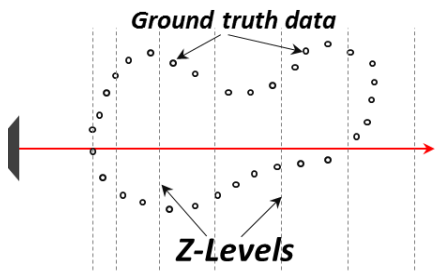

Figure 5. Experimental setup

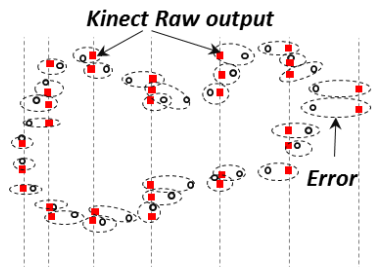

Figure 6. Kinect raw measurements

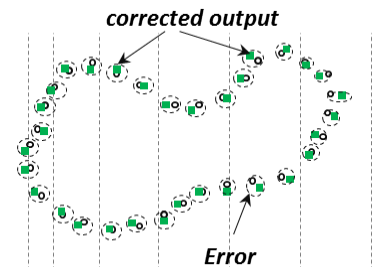

Figure 7. Corrected measurements

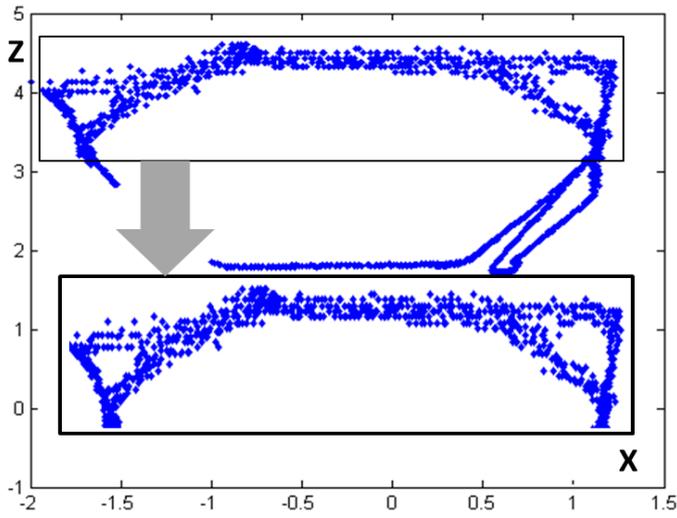

Figure 8. Raw trajectory

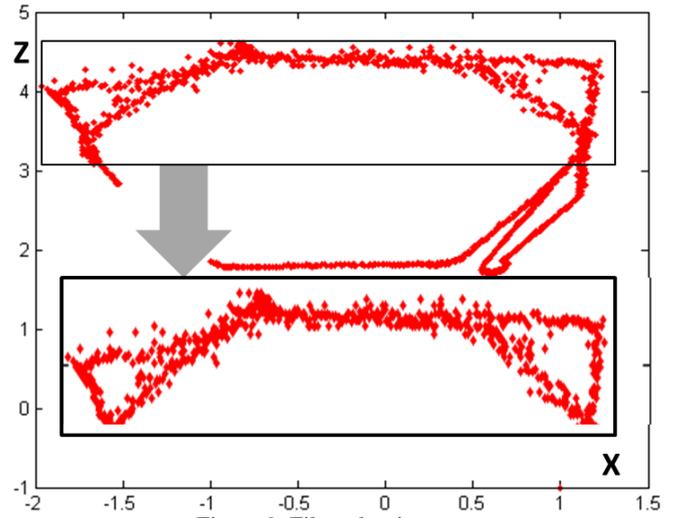

Figure 9. Filtered trajectory

$$u_{i\_rgb} = ( x'_i \ f_{x\_ir} \ / \ z'_i ) + \ c_{x\_ir} \quad (10)$$

$$v_{i\_rgb} = ( y'_i \ f_{y\_ir} \ / \ z'_i ) + \ c_{y\_ir} \quad (11)$$

All the computations of this stage are executed in parallel for every pixel on the GPU. The output is a coloured 3D point cloud where every point $P(x_i, y_i, z_i)$ has its own colour information and world coordinates.

*B. Markers extraction*

*1) RGB to HSV conversion*

RGB colour model (Figure 11) is the mostly used in computer and electronic systems such as televisions and ordinary cameras. However, this well established coding has some inaccuracies when we consider the colour from a perceptual point of view [32]. In other words, if we are facing the problem of deciding whether an object of a known colour exists in a given image in the way a human being does, we notice a large difference between the respective RGB combinations of the source and target colours, even if a small change in lighting intensity occurs. On the other hand, HSV colour coding (Figure 12). was proven to be more robust to unsteady lighting conditions and shadow [38]. Figure 13 depicts a case where we have a homogeneously coloured surface (orange). Nevertheless, the exposure to light creates differences between the textures of the different regions which are initially supposed to have the same colour (orange). The distance (L2 distance in the 3D respective colour spaces between two colour codes) in the RGB space between the two areas surrounded by the squares in Figure 13 is **50.16**. Whereas in the HSV space it is only **11.66**.

This nature of HSV coding helps enormously finding an object of a given colour in a video streams. For our specific case, HSV colour coding is useful to efficiently fetching the markers in the images we are processing.

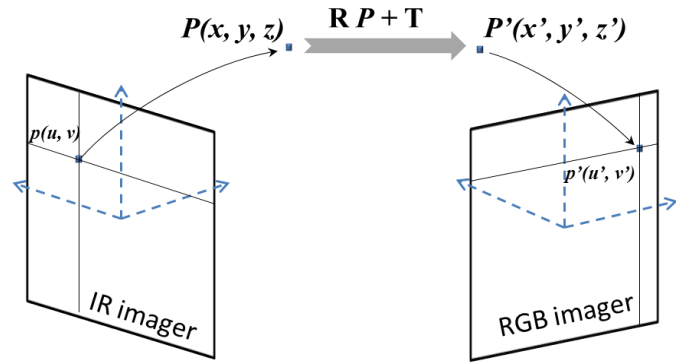

Figure 10. Kinect IR/RGB mapping

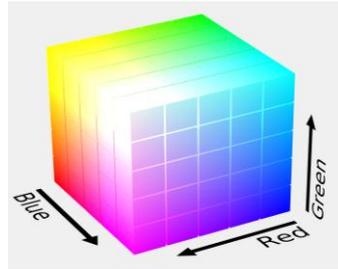
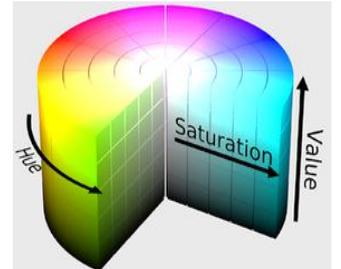

Figure 11. RGB colour space    Figure 12. HSV colours space

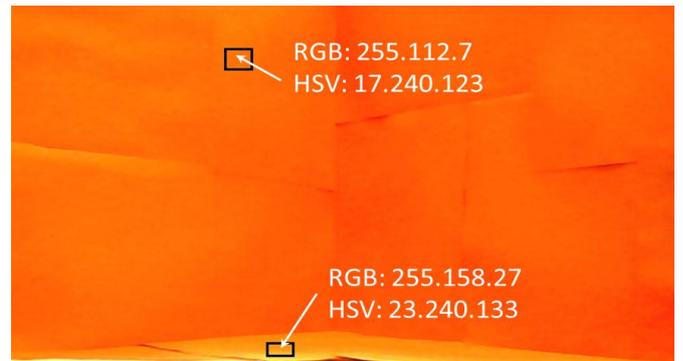

Figure 13. RGB_Diff = 50.16 ; HSV_Diff = 11.66

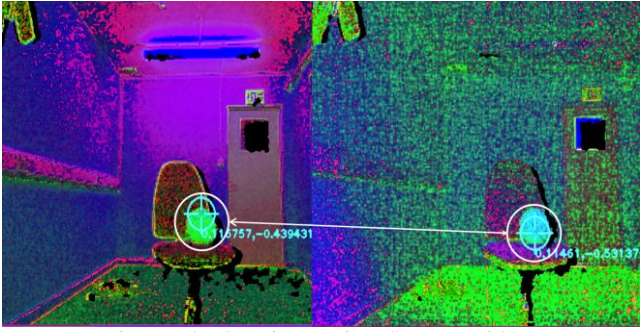
Figure 14. The Distance in HSV space 34.91

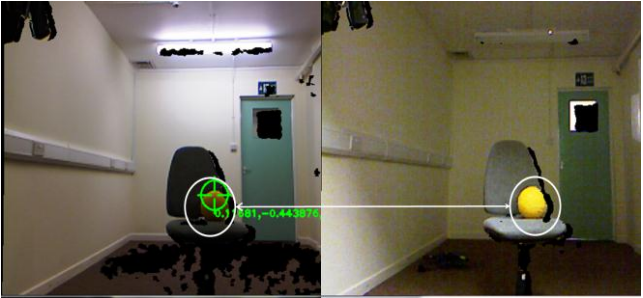
Figure 15. The Distance in RGB space 206.43

In the following example Figure 14, we have a yellow ball on top of the chair tracked in different lighting conditions (with the light of the room turned on then off). The objective of our markers extractor algorithm is to separate the markers on the robot (Figure 1) from the background.

In our experiment, the robot is assumed to move freely over different positions where luminosity is not necessarily the same. Consequently our extractor should be robust to these unstable lighting conditions to maintain an accurate tracking during the whole scenario.

Figure 14, Figure 15 illustrate the experiment where we try to fetch a yellow ball captured by Kinect with the light of the room turned on (the left image), and the image on the right captured with the light switched off. HSV images (Figure 14) are steadier as the distance between the two colours of the target in the two different images is smaller (**34.91**). The ball (in the white circle) appears clearly with almost the same HSV colour. However, with RGB images (Figure 15) the distance is **206.43**. As a consequence, the tracking cross does not appear on the right RGB image, because the algorithm was initially set to track the target whose colour was taken from the scene with the light switched on.

This example explains the need of converting Kinect's colour stream from RGB to HSV. Computationally, the complexity of conversion algorithm is linear to the size of the RGB image $O(640 \times 480)$. Moreover, this operation is scheduled on the GPU, because the same conversion kernel runs in parallel for all the pixels in the image.

*2) Colour thresholding and morphological operations*

Once we have converted our image to HSV space, we need to localise the areas which have the same colour as the target (the target colour threshold is initially captured in normal lighting conditions).

The aim of thresholding and binarisation is to recognise and extract the three yellow markers from the background. Then the alignment between the colour image and the depth map allows us to issue the 3D coordinates for each marker.

Thresholding parameters should be tuned to fit all the possible colours that can be taken by the markers in different lighting conditions during the whole capture scenario. After that, we proceed to the binarisation of the HSV image by attributing a white colour to the markers and a black colour to all the remaining parts of the frame Figure 16. The extraction of markers in the binary space facilitates the computation of the respective centres of mass for each target.

We apply erosion on the binary image to eliminate the disturbing noise spots, followed by a dilatation to then recover the eroded part of the areas representing the markers. The 3D position of the robot is the centre of the triangle composed by the three markers. To this end, all the markers should be visible to a given camera for the computation of the rigid body position. The heading of the robot is computed based on the centre of the triangle and the front marker.

*3) Markers extraction*

After localising the markers in the binary image, we proceed to the actual computation of their centres of mass. We first start by the extraction of the contours for everyone in the binary image (white spots Figure 16 (b)). Afterwards, we compute the zeroth and first moment of each marker in the binarised image [13].

Analytically the moment of a two variables function are given by:

$$\mu_{m,n} = \iint\limits_{-\infty}^{+\infty} (x - c_x)^m (y - c_y)^n f(x,y)\, dy\, dx \quad (12)$$

Here, $f(x, y)$ is the actual image and is assumed to be

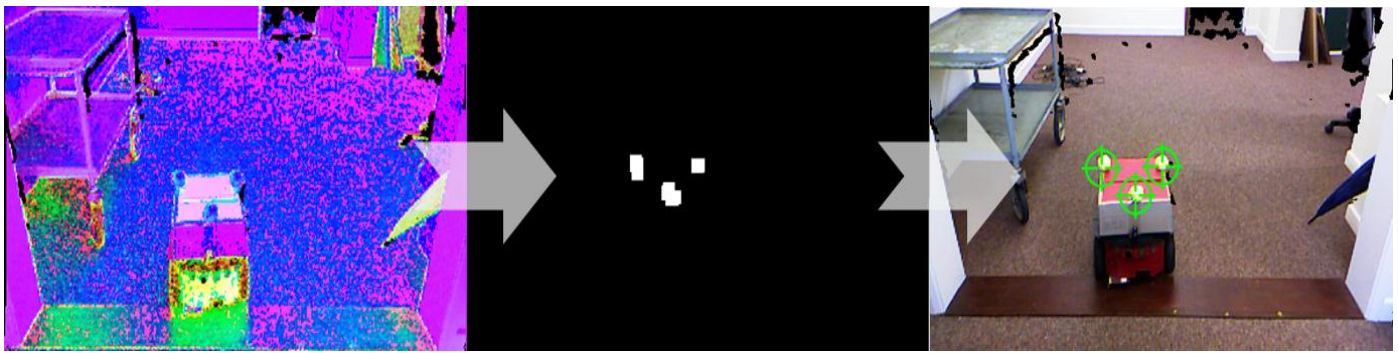

(a) HSV image          (b) binary image          (c) tracked markers
Figure 16. Markers extraction

continuous. The discrete version of (12) is:

$$\mu_{m,n} = \sum_{x=0}^{\infty} \sum_{y=0}^{\infty} (x - c_x)^m (y - c_y)^n f(x,y) \quad (13)$$

The moments are generally calculated around the origin $(c_x, c_y) = (0,0)$ as the mean value. Equation (13) becomes:

$$\mu_{m,n} = \sum_{x=0}^{\infty} \sum_{y=0}^{\infty} x^m y^n f(x,y) \quad (14)$$

To compute the centroid of the markers in our binary image, we first calculate the first moments $\mu_{01}$, $\mu_{10}$ and the zeroth moment $\mu_{00}$ (the area covered by the marker). The marker's centroid coordinates are;

$$(x_0, y_0) = (\frac{\mu_{10}}{\mu_{00}}, \frac{\mu_{01}}{\mu_{00}}) \quad (15)$$

For a binary image, the moment is the sum of the white pixels coordinates forming the contour over the area of the target marker, all in pixel metric.

This technique is robust to noise. The centroid might be a little bit shifted because of some noisy contour elements. However, the error in its position does not significantly affect the accuracy of our tracking even if the target is further away. The size of the markers is inversely proportional to the distance from the sensor. In other words, when the robot is far away from the camera, the tracked spots representing the markers in the image become smaller. Consequently, the error in their corresponding position estimation decreases. The size of the marker in the image also depends on the size of the patch we used in the morphological operations.

## V. ROBUST FILTERING

### A. Motion model

Raw Kinect position measurement is only precise at a closer range from the sensor (error in position remains below 5cms for 3 meters distance from the sensor without the correction module we introduced, and it is below 5cms for up to 6m depth after the correction stage we added). However, accurate tracking of freely moving objects requires a higher accuracy with an acceptable extra computation load. To fulfil this requirement in a relatively large indoor space, where the ground and aerial vehicles can move freely, we first correct the raw measurements of the sensor at the same time of the capture. Then we apply robust Kalman/$H_\infty$ filtering [33] on the trajectory resulting from the cameras to smooth position data of the robot. Our need to such a filtering scheme is justified by its ability to deal with the problem of uncertainty in the model describing the motion of the vehicle. If we use a stiff filter (highly tight to the parameters of the system and the disturbing noise) with an inadequate model, the tracking would fail within few iterations. Accurate tracking could not be otherwise recovered without the exact model of the system.

The tracked entities are assumed to evolve irregularly. Translation, rotation and occlusion between rigid bodies often occur in real situations. As a consequence, a constant velocity model is difficult to adapt to such a freely moving robot. On the other hand, the acceleration of ordinary ground and flying robots is more stable and does not change in amplitude like it does in sign, because of the smooth and gradual increase/decrease in velocity. Consequently, the vehicle tends to move according to a Newtonian model with varying velocity and bounded acceleration. The aim of our work is to correct the raw position data output by the camera in real time (at the frame rate of the capture). Thus, the filter should be able to robustly predict the next state of the vehicle $[x_k \ y_k \ z_k \ \dot{x}_k \ \dot{y}_k \ \dot{z}_k]^T$, and correct it accordingly after getting the measurements (the filter does not apply on orientation data, the computation of the latter is based on the filtered positions of the markers).

For $(x, y, z)$ position of a given marker, the motion model will be:

$$\begin{cases} x^+ = x + \Delta t \dot{x} + \dfrac{\Delta t^2}{2} \ddot{x} \\ y^+ = y + \Delta t \dot{y} + \dfrac{\Delta t^2}{2} \ddot{y} \\ z^+ = z + \Delta t \dot{z} + \dfrac{\Delta t^2}{2} \ddot{z} \end{cases} \quad (16)$$

The equations of the corresponding velocities $(\dot{x}, \dot{y}, \dot{z})$ are:

$$\begin{cases} \dot{x}^+ = \dot{x} + \Delta t \ddot{x} \\ \dot{y}^+ = \dot{y} + \Delta t \ddot{y} \\ \dot{z}^+ = \dot{z} + \Delta t \ddot{z} \end{cases} \quad (17)$$

The state-transition system is of the form:

$$\begin{aligned} s_{k+1} &= F s_k + B u_k + w_k \\ t_k &= H s_k + v_k \end{aligned} \quad (18)$$

Where:

$$\begin{aligned} s_k &= [x_k \ y_k \ z_k \ \dot{x}_k \ \dot{y}_k \ \dot{z}_k]^T \\ t_k &= [\check{x}_k \ \check{y}_k \ \check{z}_k] \\ u_k &= [\ddot{x}_k \ \ddot{y}_k \ \ddot{z}_k] \\ H &= [I_3, 0_3] \end{aligned} \quad (19)$$

At time-step $k$; $s_k$ is the estimated position and velocity of the vehicle; $t_k$ is the measurement output by the sensor; $u_k$: Acceleration of the vehicle; $w_k$ : Covariance of noise process affecting the system; $v_k$ : Covariance of noise process affecting the measurements.

From equations (16), (17); the state-transition matrix $F$ becomes **(20)**.

### B. Robust mixed Kalman/$H_\infty$ filter

In practical situations, the exact model of the system may not be available. The performance of such system becomes an important issue. To better explain the Kalman/$H_\infty$ filter, the authors in [18] gave the following example:

$$F = \begin{bmatrix} 1 & 0 & 0 & \Delta t & 0 & 0 \\ 0 & 1 & 0 & 0 & \Delta t & 0 \\ 0 & 0 & 1 & 0 & 0 & \Delta t \\ 0 & 0 & 0 & 1 & 0 & 0 \\ 0 & 0 & 0 & 0 & 1 & 0 \\ 0 & 0 & 0 & 0 & 0 & 1 \end{bmatrix} \quad (20)$$

$$B = \begin{bmatrix} \frac{\Delta t^2}{2} & 0 & 0 & \Delta t & 0 & 0 \\ 0 & \frac{\Delta t^2}{2} & 0 & 0 & \Delta t & 0 \\ 0 & 0 & \frac{\Delta t^2}{2} & 0 & 0 & \Delta t \end{bmatrix}^T \quad (21)$$

$$\begin{aligned} x_{k+1} &= (F_k + \Delta F_k)x_k + w_k \\ y_k &= (H_k + \Delta H_k)x_k + v_k \end{aligned} \quad (22)$$

At time-step $k$; $w_k$ and $v_k$ are uncorrelated zero-mean white noise processes with covariance matrices $Q_k$ and $R_k$ respectively. The matrices $\Delta F_k$ and $\Delta H_k$ represent the uncertainties in the system and measurements matrices. These uncertainties are assumed to be of the form:

$$\begin{bmatrix} \Delta F_k \\ \Delta H_k \end{bmatrix} = \begin{bmatrix} M_{1k} \\ M_{2k} \end{bmatrix} \Gamma_k N_k \quad (23)$$

$M_{1k}, M_{2k}$ and $N_k$ are known matrices, and $\Gamma_k$ is an unknown matrix satisfying the bond:

$$\Gamma_k^T \Gamma_k \leq I \quad (24)$$

The notation $A \leq B$ denotes that $(A - B)$ is a negative semi definite matrix. Assuming that $F_k$ is non-singular. This assumption is not too restrictive; $F_k$ should be non-singular for real systems because it comes from exponential of the system matrix (the matrix exponential is always non-singular). The problem is to design a state estimator of the form:

$$x_{k+1} = \tilde{F}_k x_k + K_k y_k \quad (25)$$

With the following characteristics:
- The estimator should be stable (the eigenvalues of $\tilde{F}_k$ are less than one in magnitude).
- Estimator error satisfies the following worst-case bound:

$$\max_{w_k, v_k} \frac{\|\tilde{x}_k\|_2}{\|w_k\|_2 + \|v_k\|_2 + \|\tilde{x}_0\|_{S_1^{-1}} + \|x_0\|_{S_2^{-1}}} \leq \frac{1}{\theta} \quad (26)$$

- The estimation error $\tilde{x}_k$ satisfies the following RMS bound:

$$E(\tilde{x}_k \tilde{x}_k^T) < P_k \quad (27)$$

The solution to the problem can be found by the following procedure:
1) Choose some scalar sequence $\alpha_k > 0$ and a small $\varepsilon > 0$.
2) Define the following matrices

$$\begin{aligned} R_{11k} &= Q_k + \alpha_k M_{1k} M_{1k}^T \\ R_{12k} &= \alpha_k M_{1k} M_{2k}^T \\ R_{22k} &= R_k + \alpha_k M_{2k} M_{2k}^T \end{aligned} \quad (28)$$

3) Initialise $P_k$ and $\tilde{P}_k$ as follows:

$$\begin{aligned} P_0 &= S_1 \\ \tilde{P}_0 &= S_2 \end{aligned} \quad (29)$$

$S_1, S_2$ are the initial values we attribute to estimation error covariance matrices for the computation of $R_{1k}, R_{2k}, F_{1k}, H_{1k}$ and $T_k$. Although these parameters have initially large values, the filter automatically tunes $P_k, \tilde{P}_k$ within few iterations. As a consequence, the process reaches a steady state, and the error in estimation decreases to its lowest levels.

4) Find positive definite solutions $P_k$ and $\tilde{P}_k$ satisfying the following Riccati equations:

$$\begin{aligned} P_{k+1} =\ & F_{1k} T_k F_{1k}^T + R_{11k} + R_{11k} R_{2k} R_{11k}^T - \\ & (F_{1k} T_k H_{1k}^T + R_{11k} R_{2k} R_{12k}) R_k^{-1} (F_{1k} T_k H_{1k}^T + R_{11k} R_{2k} R_{12k})^T \\ & + \varepsilon I \end{aligned} \quad (30)$$

$$\tilde{P}_{k+1} = F_k \tilde{P}_k F_k^T + F_k \tilde{P}_k N_k^T (\alpha_k I - N_k \tilde{P}_k N_k^T)^{-1} N_k \tilde{P}_k F_k^T + R_{11k} + \varepsilon I \quad (31)$$

Where the matrices $R_{1k}, R_{2k}, F_{1k}, H_{1k}$ and $T_k$ are defined as:

$$\begin{aligned} R_{1k} &= (\tilde{P}_k^{-1} - N_k^T N_k / \alpha_k)^{-1} F_k^T & (32) \\ R_{2k} &= R_{1k}^{-1} (\tilde{P}_k^{-1} - N_k^T N_k / \alpha_k)^{-1} R_{1k}^{-T} & (33) \\ F_{1k} &= F_k + R_{11k} R_{1k}^{-1} & (34) \\ H_{1k} &= H_k + R_{12k}^T R_{1k}^{-1} & (35) \\ T_k &= (P_k^{-1} - \theta^2 I)^{-1} & (36) \end{aligned}$$

5) If Ricatti equation solutions satisfy

$$\frac{1}{\theta^2} I > P_k \quad (37)$$

$$\alpha_k I > N_k \tilde{P}_k N_k^T \quad (38)$$

Then the estimator of equation (25) solves the problem with:

$$\begin{aligned} K_k &= (F_{1k} T_k H_{1k}^T + R_{11k} R_{2k} R_{12k}) \tilde{R}_k^{-1} & (39) \\ \tilde{R}_k &= H_{1k} T_k H_{1k}^T + R_{12k}^T R_{2k} R_{12k} + R_{22k} & (40) \end{aligned}$$

$$\hat{F}_k = F_{1k} - K_k H_{1k} \qquad (41)$$

The parameter $\varepsilon$ is generally chosen as a very small positive number. In our case it was fixed to $\varepsilon = 10^{-8}$.

The parameter $\alpha_k$ has to be chosen large enough so that the conditions of equations (37), (38) are satisfied. However, as $\alpha_k$ increases $P_k$ also increases, which results in a looser bound for the RMS estimation error [18].

A steady-state robust filter can be obtained by letting the parameter $P_{k+1} = P_k$ and $\tilde{P}_{k+1} = \tilde{P}_k$ in equations (31).

In our case, we compared the raw tracking results issued by Kinect (after Kalman filtering of the depth data) against the trajectory filtered with robust Kalman/$H_\infty$ filter. The added value of this filtering stage is a more robust tracker without a considerable delay during the relatively large amount of computation. The motion model of the robots is not initially known, hence the adaptation of robust Kalman/$H_\infty$ filtering scheme is prooved to be flexible and able to issue accurate state estimation based on the coarse system parameters. This asset enables us to track the vehicles without the exact knowledge about their motion model. Robust Kalman/$H_\infty$ filter showed very interesting results in several automation and control application [39] [40]. However, up to our knowledge we are the first who have applied it to track a moving vehicle without an exact knowledge about the state-transition model of motion. More importantly, the results we got and the error in estimation compared to the ground truth measurements show the effictivness of our approach against the naïve filtering scheme (without considering the uncertainties in the system). The latter does not have the ability to cope with the uncertainty in the parameters governing the motion of the vehicle. Nevertheless, if an exact linear model is available, robust Kalman/$H_\infty$ filter performs as efficiently as Kalman filter. Although in many real cases the exact model is hard to be determined without any error [41]. Robust filter combines the robustness of $H_\infty$ (it is less impinged by the accuracy of system's parameters and the nature of noise process affecting the measurements and the model of motion) and the optimality of Kalman filtering for linear systems. If the state-transition model of the vehicle is not linear, the adapted Robust filter still can cope with tracking by keeping the error below a predefined bond (this depends on the tuning of the filter).

Our first concern was the design of a more general tracker capable of coping with the uncertainty in the system of motion. Consequently, though the model we used to prove the effectivness of our adaptation is very simple, the results we got from a real scenario were good as we will show in results and descussion section.

## VI. TRACKING DATA FUSION

Tracking a moving vehicle in real-time could be achieved with one depth camera alone. However, Error in the measurements would be very important especially if the target evolves far from the sensor. In addition, if there are many targets moving around obstacles in the same scene, occlusions can appear and consequently prohibiting the decent recognition of the vehicles from a fix viewpoint. At this last level of our pipeline, we investigate the use of covariance intersection technique [19] to combine the estimated position outputs of all the cameras (after being filtered by robust Kalman/$H_\infty$ filter) in one consistent estimate that precisely determines a unified state of the vehicle in its space.

### A. Covariance intersection Filtering

Based on the estimated position of the vehicle issued by robust Kalman/$H_\infty$ filter $\hat{x}_{kn}$ ($1 \leq n \leq N$) at time step $k$, and the corresponding positive error covariance matrices $P_{kn}$; we compute a combined estimate $\tilde{x}$ with its error covariance $P$ where the true state of the system is $x$ (real position of the vehicle).

If we consider $N$ unbiased estimates related to each camera $\hat{x}_1, \hat{x}_2, \hat{x}_3 \ldots \hat{x}_N$ for the unknown state vector $\tilde{x}$:

$$\tilde{x} = P \sum_{n=1}^{N} P_n^{-1} \hat{x}_n \qquad (42)$$

$$P^{-1} = \sum_{n=1}^{N} P_n^{-1} \qquad (43)$$

In the presence of correlation between estimation errors, the estimated $P$ may become far too optimistic and this can cause divergence in sequential filtering. A conservative estimate can be given by applying covariance intersection according to:

$$\tilde{x} = P \sum_{n=1}^{N} \omega_n P_n^{-1} \hat{x}_n \qquad (44)$$

$$P^{-1} = \sum_{n=1}^{N} \omega_n P_n^{-1} \qquad (45)$$

With the nonnegative coefficients $\omega_n$ verifying the following consistency condition:

$$\sum_{n=1}^{N} \omega_n = 1 \qquad (46)$$

An estimate could always be obtained with:

$$P \geq P_0 := E[(\tilde{x} - x)(\tilde{x} - x)^T] \qquad (47)$$

Where $P \geq P_0$ denotes the fact that $P - P_0$ is positive semi-definite. Consequently, the coefficients $\omega_n$ are meant to minimise either the trace or the determinant of $P$.

In order to avoid the possibly high numerical implementation effort for finding the solution of such a highly nonlinear optimisation problem, Neihsen [19] has proposed a fast approximate solution instead. Hence, for $trace(P_n) \leq trace(P_m); 1 \leq n, m \leq N$ one would expect $\omega_n \geq \omega_m$.

From a computationally optimal point of view, rather than using estimation uncertainty $P_n$ the authors in [34] introduced estimation certainty by considering $I_n = P_n^{-1}$ :

$$\omega_n = \frac{trace(I_n)}{\sum_{i=0}^{N} trace(I_i)} \qquad (48)$$

Equation (48) means that the greater is $trace(I_n)$ (the more certain we are about the estimate $\hat{x}_n$), the higher is the corresponding weight $\omega_n$. On the other hand, the smaller is $trace(I_n)$ the lower is the weight $\omega_n$. More importantly; consistency condition (46) remains satisfied:

$$\sum_{n=1}^{N} \omega_n = \frac{\sum_{n=0}^{N} trace(I_n)}{\sum_{i=0}^{N} trace(I_i)} = 1 \qquad (49)$$

### B. Covariance intersection for multikinect tracker

As we explained earlier in section III.A.2), some Kinects may be faulty and consequently produce erroneous measurements when the target is far away from the sensor. However, with a multiview setup the final estimate of position and orientation can be jointly corrected. The correction is led by the weighting coefficients which weigh better the more accurate measure that could be delivered by each of the cameras constituting the multiview setup which covers the whole scene.

Another contribution of the present work, is the adaptive weighting scheme based not only on the assessment of quality for each estimate resulting from robust Kalman/$H_\infty$ filter using estimate covariance, but also from the confidence in the raw measure issued by the camera itself.

The idea is based upon the introduction of a quality factor for each of the cameras capturing the motion of the vehicle. This quality indicator is issued from the remaining error in the sensor after applying the appropriate correction model.

The mathematical formulation is as follows:

For the processing thread of the *n*-th camera:

$P_n$ : is the covariance matrix of the error in the estimate issued by robust Kalman/$H_\infty$ filter.

$K_n$ : is the covariance matrix characterising the residue error after correction.

$Z_n$ : is a positive scalar factor representing the distance between the target and the camera. This latter assumption is motivated by the fact that the smaller is the depth of the target the more accurate is the measurement of position as we explained in section III.A.3). The accuracy of the sensor is inversely proportional to the depth value.

Figure 17 depicts the situation where every sensor has its native hardware accuracy matrix $K_n$ (Red circles). Moreover, depending on the distance separating the camera from the target ($Z_n$), we introduce the weighting coefficient $Z_n$:

For every pair of position estimates $\hat{x}_n, \hat{x}_m$ this condition should be satisfied:

$tr(K_n) + tr(P_n) + Z_n \leq tr(K_m) + tr(P_m) + Z_m \Rightarrow \omega_n \geq \omega_m ;$ (50)
$1 \leq n, m \leq N$

Equation (50) means that $\hat{x}_n$ affects the final estimate $\tilde{x}$ more than $\hat{x}_m$ does. And $P_n$ affects the final error in estimation $P$ more than $P_m$ does. In our tracking algorithm, we considered Niehsen's finding about fast covariance intersection. In addition, we included the uncertainty characterising the quality of measure issued by the sensor. Our weighting coefficients are given by the expression below:

$$\omega_n = \frac{\sum_{\substack{i=1 \\ i \neq n}}^{N} (tr(K_i) + tr(P_i) + Z_i)}{\sum_{i=1}^{N} (tr(K_i) + tr(P_i) + Z_i)} \qquad (51)$$

Another form of the same expression is more adequate to reduce the load of computation:

$$\omega_n = \frac{\sum_{i=1}^{N}(tr(K_i) + tr(P_i) + Z_i) - (tr(K_n) + tr(P_n) + Z_n)}{\sum_{i=1}^{N}(tr(K_i) + tr(P_i) + Z_i)} \qquad (52)$$

Consequently, we only need to compute the traces of the matrices once. The denominator $\sum_{i=1}^{N}(tr(K_i) + tr(P_i) + Z_i)$ is also computed once. Afterwards, we just subtract the corresponding parameter $(tr(K_n) + tr(P_n) + Z_n)$ appropriate to every one of the estimates.

The condition of consistency (46) remains verified as $\sum_{n=1}^{N} \omega_n = 1$. Experiments conducted using this approach proved formula (52) to be more realistic and suitable for real tracking scenarios as will show the results we got. The time taken to compute the traces of the matrices ($3 \times 3$ matrices), does not significantly affect the overall performance of the system.

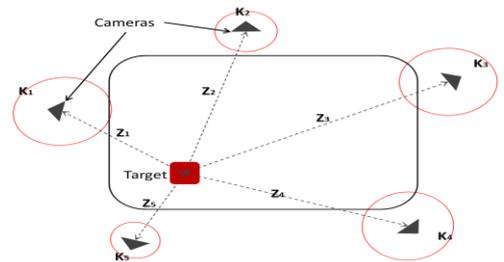

Figure 17. Covariance intersection parameters

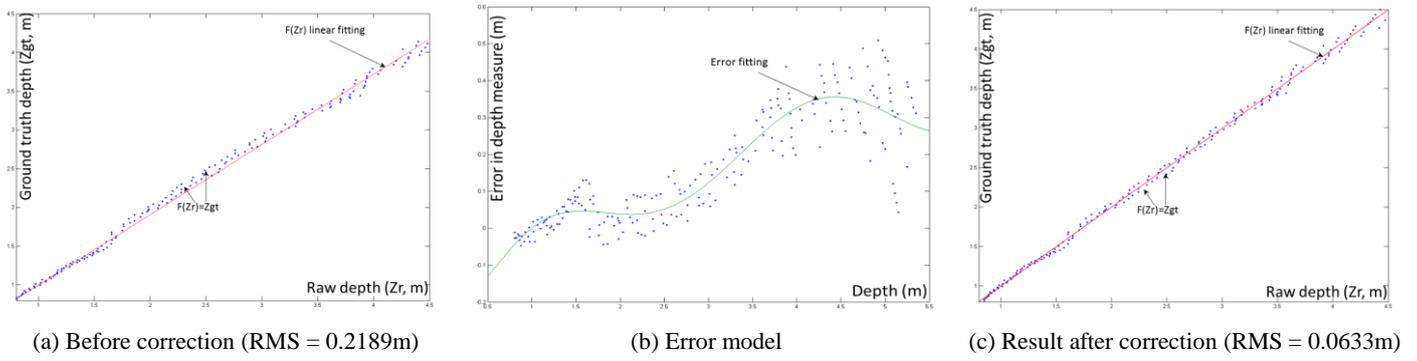

(a) Before correction (RMS = 0.2189m)  (b) Error model  (c) Result after correction (RMS = 0.0633m)

Figure 18. Drifty kinect

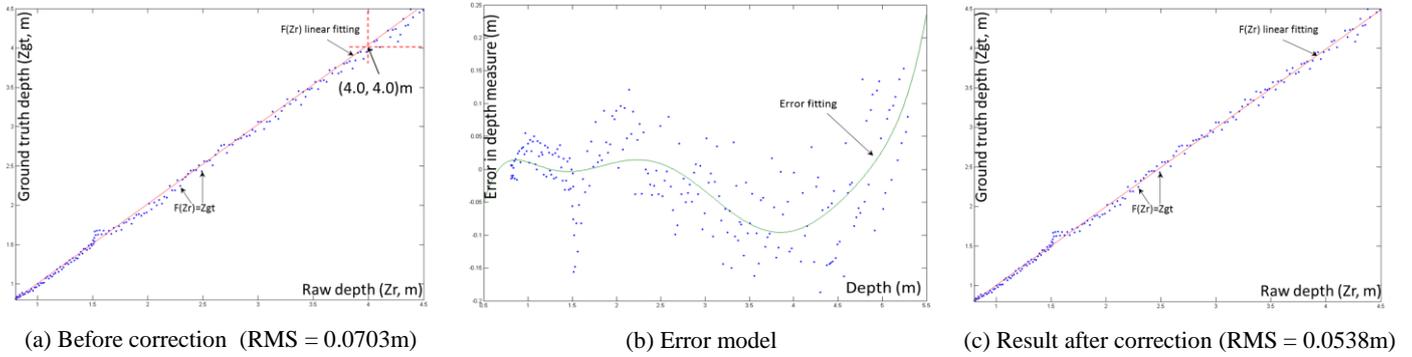

(a) Before correction (RMS = 0.0703m)  (b) Error model  (c) Result after correction (RMS = 0.0538m)

Figure 19. Good kinect

## VII. RESULTS AND DISCUSSION

### A. Sensor first stage correction:

To illustrate the effect of our sensor correction method, we conducted some experiments during which two Kinects of different measuring precisions were corrected. Figure 18, Figure 19 illustrate two cases where a drifty and a sane sensor were adjusted using an eight degree polynomial. The first column of both figures ((a) before correction) holds the graphs of the function $f(z_{sh}) = z_{cor}$. ($z_{sh}$: shifted depth; $z_{cor}$: ground truth reference depth reading).The more the fitting line approaches $y = x$, the more accurate our sensor will be (the range measure output by Kinect is closer to the ground truth). With the faulty sensor (Figure 18), the curve representing $f(z_{sh})$ is clearly shifted below $y = x$. This happens because the sensor overestimates the range of objects in front of it. As a consequence, such behaviour limits the its ability to fully capture the 3D geometry within the working range (0.8m-4.0m). On the other hand, with the good sensor (Figure 19), the representative curve is almost superposed on $y = x$ when the depth of the object is below 4.0m. After we get further than permitted by the manufacturer ($z_{sh} > 4.0m$), the accuracy drops and the corresponding fitting line shifts above $y = x$.

To correct the sensor, we plot the points $(z_{sh}, z_{sh} - z_{cor})$. We then fit them with a polynomial $f(z_{sh})$ that minimises the error between $z_{sh}$ and $z_{cor}$) in the least squares sense.

In practice, we found that an eight degree polynomial represents well the set of points with a small error factor. The correct depth of the shifted one ($z_{sh}$) is obtained by evaluating $f(z_{sh})$ for the whole depth map.

Because of the limited number of discrete depth values that can exist in any point cloud issued by Kinect, we attribute to each raw range value ($Z_{sh}$) a respective corrected value ($Z_{cor}$). As a result, the correction of the depth image is reduced to the correction of the known depth levels [37].

This module (Depth correction) runs on the GPU and it is pixel-wise. We apply the correction on every valid depth reading before any further processing to ensure a better quality of data. This allows us to fully benefit from the available accuracy of the sensor. After the correction (column (c) in Figure 18, Figure 19), the depth data is almost the same as the ground truth one. However, for larger values the resolution of the sensor decreases and only some discrete measurements could be obtained. As we can see from the figures, the density of the samples we used to compute the correction model decreases with increasing range.**TABLE 2** shows the error in measurements for some Kinects used in our multiview tracking experiments:

|  | Kinect_0 | Kinect_1 | Kinect_3 | Kinect_4 |
|---|---|---|---|---|
| Before | 0.1114 | 0.1474 | 0.2189 | 0.0703 |
| After | 0.0490 | 0.0598 | 0.0633 | 0.0538 |

TABLE 2. RMS error in meter, for some Kinects before and after applying the corresponding correction model.

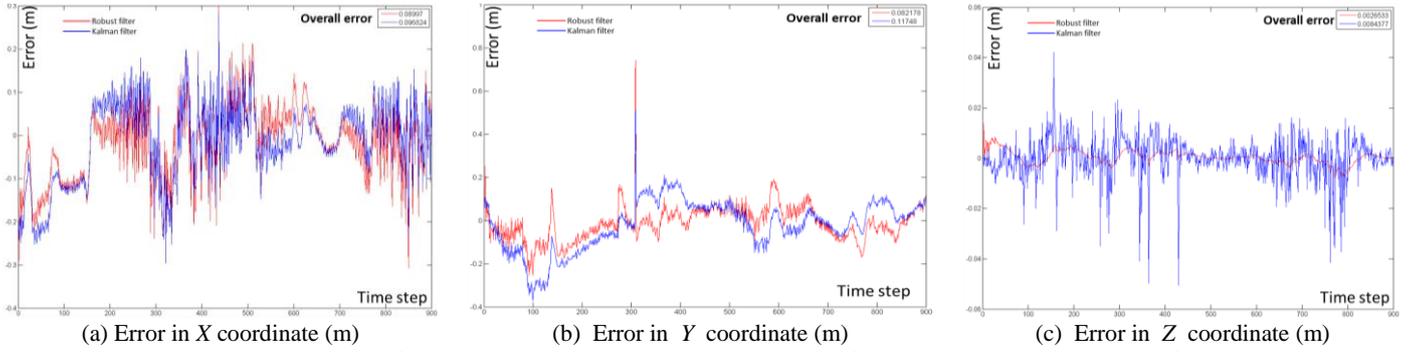

(a) Error in *X* coordinate (m)  (b) Error in *Y* coordinate (m)  (c) Error in *Z* coordinate (m)

Figure 20. The best tracking results after applying Robust Kalman/$H_\infty$ and Kalman filters on *X,Y* and *Z* coordinates of the robot over time

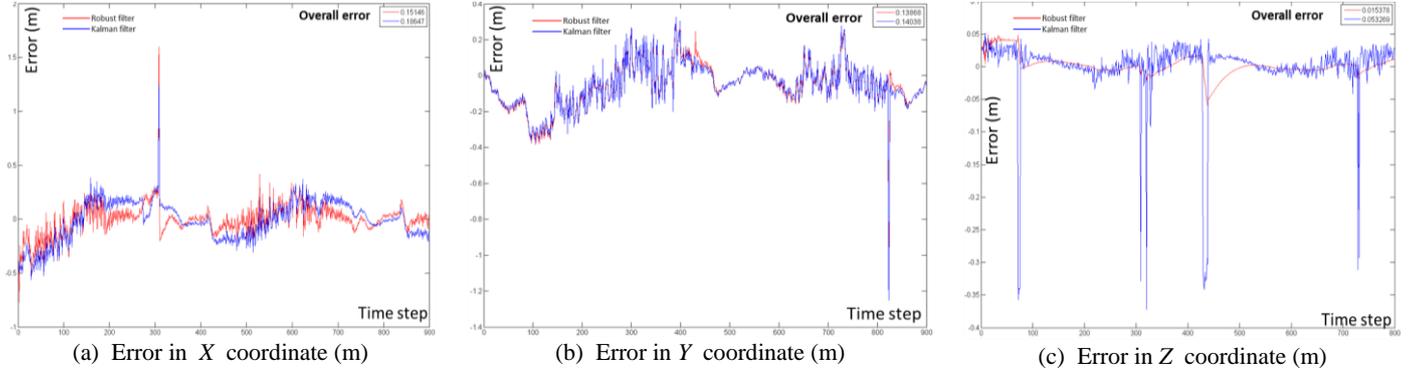

(a) Error in *X* coordinate (m)  (b) Error in *Y* coordinate (m)  (c) Error in *Z* coordinate (m)

Figure 21. The worst tracking results after applying Robust Kalman/$H_\infty$ and Kalman filters on X,Y and Z coordinates of the robot over time

*B. Robust Kalman/$H_\infty$ filter*

The results discussed in the following sections are issued from trajectories similar to the one in Figure 23.

Figure 20, Figure 21 show respectively the best and the worst performance of our Robust Kalman/$H_\infty$ (**RF**) tracking algorithm along with comparative results when using a classical Kalman filter (**KF**) (this was possible with Kalman filter because the modelled system is linear) .As we explained earlier, and for sake of generality in our tracking algorithm, we assumed that the exact state/transition model of the robots is unknown and we approached it with a generic Newtonian system. To furthermore overcome the uncertainty in the model, we adapted Robust Kalman/$H_\infty$ filter which has the ability to deal with imprecise parameters.

For *X* and *Y* coordinates (*X*, *Y* and *Z* axes are shown in Figure 22) in the two figures (Figure 20, Figure 21), the behaviour (the shape of the graphs) of the error for the two filters (Kalman and Robust Kalman/$H_\infty$) is almost similar because the model of motion is the same for both algorithms. Nevertheless, tracking error with Robust Kalman/$H_\infty$ is smaller. The smallest error within all the five cameras for *X* coordinates was **0.090m** with RF against **0.097m** for KF Figure 20 (a). The worst case in *X* was **0.15m** for RF against **0.19m** with KF Figure 21 (a). For *Y* coordinates the best results was **0.082m** with RF against **0.117m** for KF Figure 20 (b). The worst case in *Y* was **0.138m** with RF against **0.140m** for KF Figure 21 (b).

Among the three coordinates of the moving vehicles, *Z* suffers the least from noise (it slightly changes value during the scenario). The best *Z* results were **0.0027m** with robust filter against **0.0084m** for Kalman filter Figure 20 (c). The worst case for *Z* was **0.0154m** with RF against **0.0532m** for KF Figure 21 (c).

Throughout all the experiments we conducted, RF filter is less affected by the inaccuracies in the parameters of the system. More importantly, it was able to predict the position of the moving robot even if no measurements were available.

The detailed results are given in **TABLE 3**, **TABLE 4**, **TABLE 5** and **TABLE 6**. For all the sensors, RF always gives the best estimation.

The results given in these tables value in same way all the five cameras as all of them were previously corrected with their appropriate models. On the other hand, the effectiveness of some sensors against others is highly biased by the trajectory of the vehicle. To balance the contribution of all the cameras covering the scene, we configured the environment surrounding the robot in a way that enforces it to follow a subcircular motion, so it can approach each camera (this is not a requirement in our system, we just do that to show the effect of the distance from the sensor on our multiview tracking process). If all the cameras have the same accuracy, the closest one to the robot will be the best candidate able to precisely capture the position. Besides, the remaining cameras issue position data with less precision as they are further away from the vehicle Figure 17.

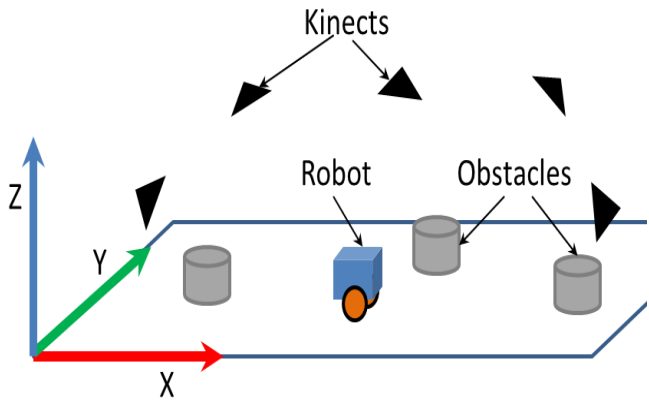

Figure 22. X,Y and Z axes in our experimental setup

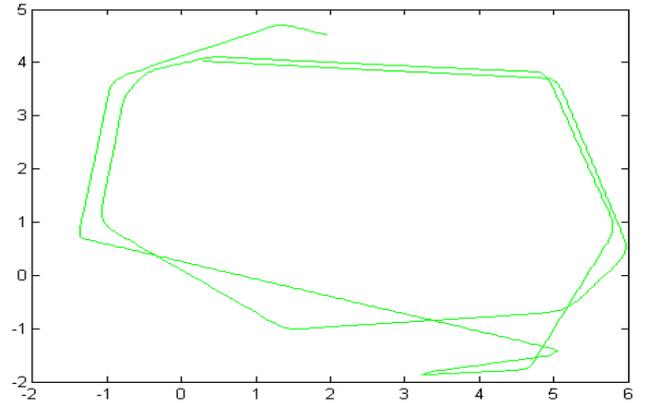

Figure 23. the ground truth trajectory followed by the robot

| X | Kinect_0 | Kinect_1 | Kinect_2 | Kinect_3 | Kinect_4 |
|---|---|---|---|---|---|
| Raw data RMS(m) | 0.1233 | 0.1897 | 0.1353 | 0.1322 | 0.0988 |
| Robust filter RMS(m) | **0.0954** | **0.1515** | **0.1266** | **0.1170** | **0.0900** |
| Kalman filter RMS(m) | 0.1207 | 0.1865 | 0.1332 | 0.1289 | 0.0968 |
| TABLE 3. Error in X component (compared to ground truth data) | | | | | |

| Y | Kinect_0 | Kinect_1 | Kinect_2 | Kinect_3 | Kinect_4 |
|---|---|---|---|---|---|
| Raw data RMS(m) | 0.1405 | 0.1199 | 0.1420 | 0.0906 | 0.1140 |
| Robust filter RMS(m) | **0.1387** | **0.0822** | **0.1360** | **0.0849** | **0.0952** |
| Kalman filter RMS(m) | 0.1404 | 0.1175 | 0.1415 | 0.0889 | 0.1123 |
| TABLE 4. Error in Y component (compared to ground truth data) | | | | | |

| Z | Kinect_0 | Kinect_1 | Kinect_2 | Kinect_3 | Kinect_4 |
|---|---|---|---|---|---|
| Raw data RMS(m) | 0.0084 | 0.0291 | 0.0316 | 0.0523 | 0.0319 |
| Robust filter RMS(m) | **0.0027** | **0.0104** | **0.0127** | **0.0154** | **0.0072** |
| Kalman filter RMS(m) | 0.0084 | 0.0294 | 0.0321 | 0.0533 | 0.0320 |
| TABLE 5. Error in Z component (compared to ground truth data) | | | | | |

| **Overall error (X,Y,Z)** | Kinect_0 | Kinect_1 | Kinect_2 | Kinect_3 | Kinect_4 |
|---|---|---|---|---|---|
| Raw data RMS(m) | 0.187119 | 0.226294 | 0.198667 | 0.168584 | 0.154192 |
| Robust filter RMS(m) | **0.168363** | **0.172677** | **0.186239** | **0.145376** | **0.131205** |
| Kalman filter RMS(m) | 0.185341 | 0.22238 | 0.196964 | 0.165406 | 0.151676 |
| TABLE 6. Overall error (compared to ground truth data) | | | | | |

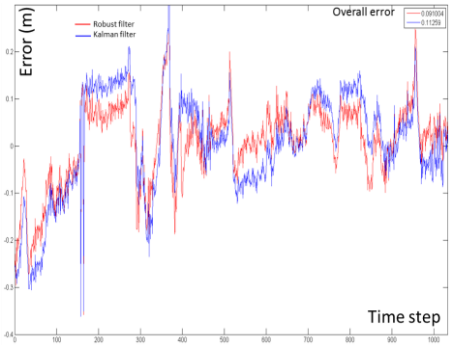
(a) Error in *X* coordinate (m)

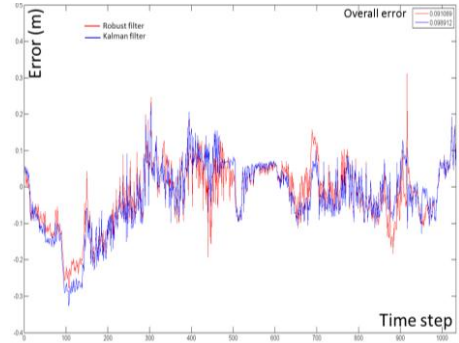
(b) Error in *X* coordinate (m)

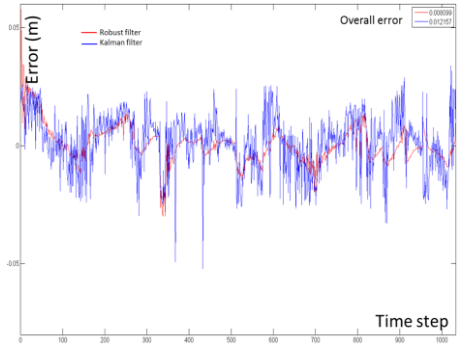
(c) Error in *Z* coordinate (m)

Figure 24. $P_n$ based weighting results

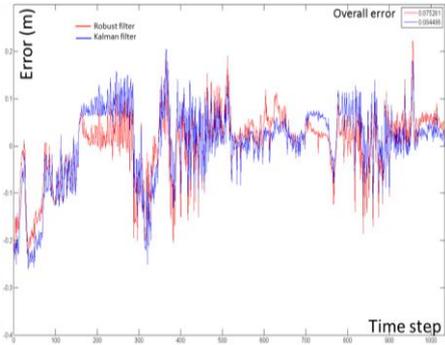
(a) Error in *X* coordinate (m)

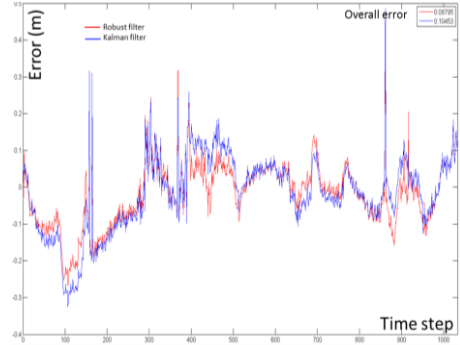
(b) Error in *Y* coordinate (m)

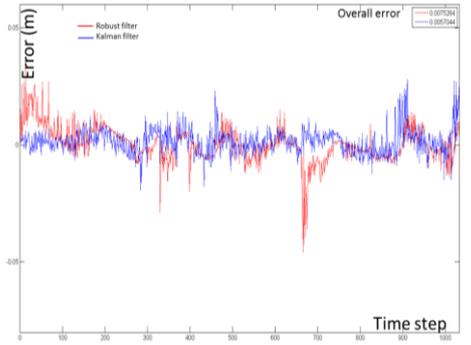
(c) Error in *Z* coordinate (m)

Figure 25. $P_n$ and $K_n$ weighting results

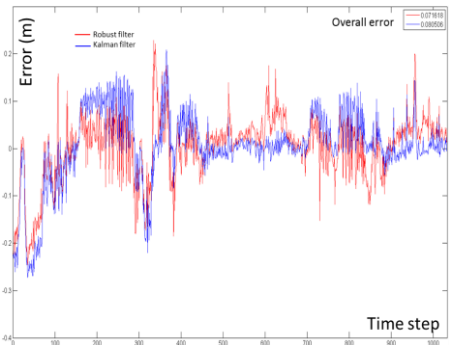
(a) Error in *X* coordinate (m)

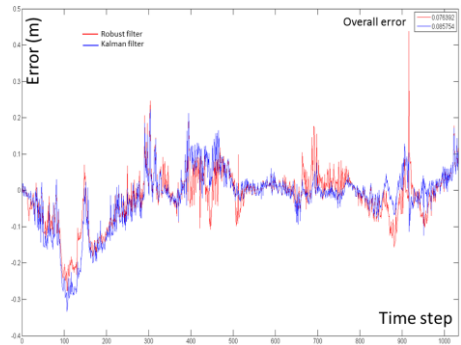
(b) Error in *Z* coordinate (m)

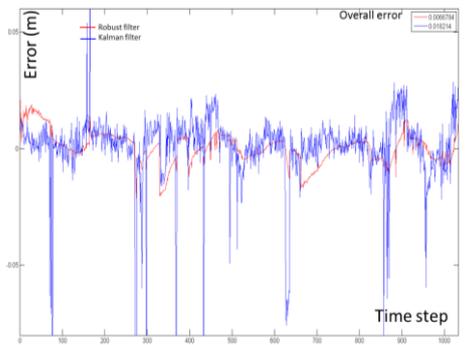
(c) Error in *Z* coordinate (m)

Figure 26. $P_n$, $K_n$ and $Z_n$ weighting results

| filters | X | Y | Z | Overall (X,Y,Z) |
|---|---|---|---|---|
| CI + RF | 0.091 | 0.091 | 0.008 | 0.128 |
| CI + KF | 0.112 | 0.098 | 0.012 | 0.149 |

Table 7. Final tracking error after **CI** filtering with $P_n$ weighting

| filters | X | Y | Z | Overall (X,Y,Z) |
|---|---|---|---|---|
| CI + RF | 0.075 | 0.087 | 0.007 | 0.115 |
| CI + KF | 0.084 | 0.104 | 0.005 | 0.133 |

Table 8. Final tracking error after **CI** filtering with $P_n$ **and** $K_n$ weighting

| filters | X | Y | Z | Overall (X,Y,Z) |
|---|---|---|---|---|
| CI + RF | 0.071 | 0.076 | 0.006 | 0.104 |
| CI + KF | 0.080 | 0.085 | 0.018 | 0.118 |

Table 9. Final tracking error after **CI** filtering with $P_n$, $K_n$ and $Z_n$ weighting

## C. Covariance intersection (CI)

At the final stage of our multiview tracking pipeline, we use covariance intersection filter (**CI**) to fuse position data resulting from sensor-wise estimates. To validate our finding about CI weighting coefficients, we compared three different approaches to apply the weights on the estimates. We tested the weighting with only the error in estimation ($P_n$) issued by RF. Then we combined the latter with the uncertainty in the accuracy of the sensor ($P_n$ and $K_n$). Finally, we combined both elements with the confidence in the depth measure ($P_n, K_n$ and $Z_n$). After considering each new parameter affecting the process of data fusion, the quality of estimation is better improved.

We tested our CI algorithm on both: the estimate of trajectory issued by Kalman filter tracking (**CI + KF**) and the one issued by Robust Kalman/$H_\infty$ filter tracking (**CI + RF**). The results of the fused trajectories were as follows:

Firstly, for $X$ coordinate $P_n$ on its own (Figure 24.(a),Table 7(col X)) gave us an error of **0.091m** with RF, whereas with KF it gives **0.112m**. After considering the accuracy of the sensor (Figure 25.(a),Table 8 (col X) ), the error was reduced by more than 2cm for both filters to reach **0.075m** with RF, and **0.084m** for KF. The introduction of $Z_n$ (Figure 26.(a), Table 9 (col X)) further approached the estimation to its ground truth counterpart it reached **0.071**m with RF and **0.080m** for KF.

Secondly, for $Y$ coordinate $P_n$ (Figure 24.(b), Table 7(col Y)) on its own gave us again an error of **0.091m** for RF, whereas with KF it gives **0.098m**. After adding the accuracy parameter of the sensor, the error was slightly reduced to **0.087m** with RF. However, the introduction of $K_n$ (Figure 25.(b), Table 8 (col Y)) adversely affected the accuracy of KF estimates as it decreased to **0.104m**. Lastly, $Z_n$ (Figure 26.(b), Table 9 (col Y)) error in estimation dropped to **0.076m**, at the same time the error with KF dropped to **0.085m**.

Thirdly, for $Z$ component, $P_n$ (Figure 24.(c), Table 7(col Z)) on its own gave us again an error of **0.008m** for RF, whereas with KF it gives **0.012m**. After adding the accuracy parameter of the sensor, the error was slightly reduced to **0.0075m** with RF. However, the introduction of $K_n$ (Figure 25.(c), Table 8 (col Z)) worsens the accuracy of KF estimates as it decreased to **0.005m**. $Z_n$ (Figure 26.(c), Table 9 (col Z)) positively affected the error for RF which was **0.006m**, the error with KF dropped to **0.018m**.

The overall tracking error with RF+CI and all the parameters considered ($P_n, K_n$ and $Z_n$) in the computation of weighting coefficients was **0.104m.** On the other hand, for KF+CI the best result was **0.118m**, it was acquired with weights computation based on all three parameters.

From the last result we got, the quality of estimation between RF and KF is not significant, this is true because of the compensating effect of CI algorithm through all the sensors. An optimistic approach would be the selection of the closest sensor to the robot at every time step. However, the accuracy of the chosen sensors is not the always the best.

Difference in the performance among cameras justifies the need to combine all the available data for a more accurate tracking. In addition, occlusions may prohibit the closest camera from capturing the vehicle. In this case, the sensor-wise filter will predict the next move based on the last state and the system's model which is basically imprecise.

## VIII. CONCLUSION AND FUTURE WORKS

In this work, we presented a novel approach to accurately track moving vehicles in indoor environments with multiple cheap consumer RGBD cameras. We explained all the details about our contributions and the adaptation of the different tools from the literature to our specific case of study. Our findings about RGBD sensor correction are of importance for a more accurate measurement with any type of sensors suffering from the same drawbacks. Up to our knowledge, we were the first who investigated and applied robust filter on objects tracking in RGBD capture. We furthermore demonstrated the power of the latter to overcome the lack of knowledge about the system governing the behaviour of the vehicles. However this filter (RF) could be applied to other real world scenarios where the system could not be exactly modelled. We considered the quality of measurements and estimates issued by each sensor in CI algorithm. We successfully combined all the single contributions of the cameras in one consistent estimate. Test results show the performance we reached at a compelling frame rate of 25Fps with five Knects. Our solution could be easily adapted to any indoor tracking scenario. For multiple targets an object association algorithm is needed though. Extra cameras in the system yield extra burden of computations, thus the frame rate is more likely to decrease. Moreover, as RGBD sensors are basically active devices, interference hinders appropriate localisation of markers in the depth map. Thus the loss of the 3D position by some but not necessarily all the sensors. The incorporation of the GPU in all the stages of processing (raw depth data correction, Kalman filtering applied on the depth map all the pixels for every camera. As well as Robust filter applied on a single position of the vehicle). For covariance intersection no parallelisation is needed (the fusion in centrally performed) because the CPU overtakes the GPU in linear execution.

As a future work, we aim to overcome interference problem by algorithmic means using the stereo RGB/IR images to complete the missing depth information. We further aim to apply the same filtering scheme on 3D reconstruction applications for object recognition purposes. Another planned work is the combination of motion and 3D structure all together for a complete acquisition of the shape and behaviour of the robots evolving in the scene in 6DOF.

REFERENCES

[1] H. Yang, L. Shao, F. Zheng, L. Wang, and Z. Song, "Recent advances and trends in visual tracking: A review," *Neurocomputing*, vol. 74, no. 18, pp. 3823–3831, Nov. 2011.

[2] K. Litomisky, "Consumer RGB-D Cameras and their Applications," 2012.

[3] J. Fung and S. Mann, "OpenVIDIA: parallel GPU computer vision," *Proc. 13th Annu. ACM Int. …*, 2005.

[4] A. Amamra and N. Aouf, "Real-Time Robust Tracking with Commodity RGBD Camera," *… , Man, Cybern. (SMC), 2013 IEEE …*, 2013.

[5] A. Yilmaz, O. Javed, and M. Shah, "Object tracking," *ACM Comput. Surv.*, vol. 38, no. 4, p. 13–es, Dec. 2006.

[6] M. Taj and A. Cavallaro, "Multi-view multi-object detection and tracking," *Comput. Vis.*, 2010.

[7] F. Lv, T. Zhao, and R. Nevatia, "Self-calibration of a camera from video of a walking human," *… 2002. Proceedings. 16th Int. …*, 2002.

[8] K. Okuma, A. Taleghani, and N. De Freitas, "A boosted particle filter: Multitarget detection and tracking," *Comput. Vision-ECCV …*, 2004.

[9] B. Leibe, E. Seemann, and B. Schiele, "Pedestrian detection in crowded scenes," *Comput. Vis. Pattern …*, 2005.

[10] A. Bobick, S. Intille, and J. Davis, "The KidsRoom: A perceptually-based interactive and immersive story environment," *Presence …*, 1996.

[11] S. Intille, J. Davis, and A. Bobick, "Real-time closed-world tracking," *Comput. Vis. Pattern …*, 1997.

[12] C. Wren and A. Azarbayejani, "Pfinder: Real-time tracking of the human body," *Pattern Anal. …*, 1997.

[13] J. Flusser and T. Suk, "Rotation Moment Invariants for Recognition of Symmetric Objects," *IEEE Trans. Image Process.*, vol. 15, no. 12, pp. 3784–3790, Dec. 2006.

[14] S. Y. Chen, "Kalman Filter for Robot Vision: A Survey," *IEEE Trans. Ind. Electron.*, vol. 59, no. 11, pp. 4409–4420, Nov. 2012.

[15] B. Merven, F. Nicolls, and G. De Jager, "Multi-camera person tracking using an extended kalman filter," *Fifteenth Annu. Symp. …*, 2003.

[16] Y. Rui and Y. Chen, "Better proposal distributions: Object tracking using unscented particle filter," *… Vis. Pattern Recognition, 2001. CVPR …*, 2001.

[17] M. Pupilli and A. Calway, "Real-Time Camera Tracking Using a Particle Filter.," *BMVC*, 2005.

[18] D. Simon, *Optimal State Estimation: Kalman, H Infinity, and Nonlinear Approaches*. Wiley-Interscience, 2006, p. 552.

[19] W. Niehsen, "Information fusion based on fast covariance intersection filtering," in *Proceedings of the Fifth International Conference on Information Fusion. FUSION 2002. (IEEE Cat.No.02EX5997)*, 2002, vol. 2, pp. 901–904.

[20] D. Smith and S. Singh, "Approaches to Multisensor Data Fusion in Target Tracking: A Survey," *IEEE Trans. Knowl. Data Eng.*, vol. 18, no. 12, pp. 1696–1710, Dec. 2006.

[21] J. Shotton, T. Sharp, and A. Kipman, "Real-time human pose recognition in parts from single depth images," *Commun. …*, 2013.

[22] D. Correa and D. Sciotti, "Mobile robots navigation in indoor environments using kinect sensor," *Crit. Embed. …*, 2012.

[23] P. Henry, M. Krainin, and E. Herbst, "RGB-D mapping: Using Kinect-style depth cameras for dense 3D modeling of indoor environments," *… J. Robot. …*, 2012.

[24] T. Nakamura, "Real-time 3-D object tracking using Kinect sensor," in *2011 IEEE International Conference on Robotics and Biomimetics*, 2011, pp. 784–788.

[25] S. Izadi, D. Kim, and O. Hilliges, "KinectFusion: real-time 3D reconstruction and interaction using a moving depth camera," *Proc. 24th …*, 2011.

[26] J. Tong, J. Zhou, L. Liu, Z. Pan, and H. Yan, "Scanning 3D full human bodies using Kinects.," *IEEE Trans. Vis. Comput. Graph.*, vol. 18, no. 4, pp. 643–50, Apr. 2012.

[27] J. Han, L. Shao, D. Xu, and J. Shotton, "Enhanced computer vision with Microsoft Kinect sensor: a review.," *IEEE Trans. Cybern.*, vol. 43, no. 5, pp. 1318–34, Oct. 2013.

[28] "kinect_calibration/technical - ROS Wiki." [Online]. Available: http://wiki.ros.org/kinect_calibration/technical. [Accessed: 27-Jan-2014].

[29] R. Reyes, I. Lopez, J. J. Fumero, and F. de Sande, "accULL: An User-directed Approach to Heterogeneous Programming," in *2012 IEEE 10th International Symposium on Parallel and Distributed Processing with Applications*, 2012, pp. 654–661.

[30] M. Andersen, T. Jensen, and P. Lisouski, "Kinect depth sensor evaluation for computer vision applications," 2012.

[31] R. Kalman, "A new approach to linear filtering and prediction problems," *J. basic Eng.*, vol. 82, no. Series D, pp. 35–45, 1960.

[32] M. Sedláček, "Evaluation of RGB and HSV Models in Human Faces Detection. Central European Seminar on Computer Graphics, Budmerice."

[33] Y. S. Hung and F. Yang, "Robust H∞ filtering with error variance constraints for discrete time-varying systems with uncertainty," *Automatica*, vol. 39, no. 7. pp. 1185–1194, 2003.

[34] D. Franken and A. Hupper, "Improved fast covariance intersection for distributed data fusion," in *2005 7th International Conference on Information Fusion*, 2005, vol. 1, p. 7 pp.

[35] J. Fu, S. Wang, Y. Lu, S. Li, and W. Zeng, "Kinect-like depth denoising," *Circuits Syst. ( …*, 2012.

[36] K. Khoshelham and S. O. Elberink, "Accuracy and resolution of Kinect depth data for indoor mapping applications.," *Sensors (Basel).*, vol. 12, no. 2, pp. 1437–54, Jan. 2012.

[37] A. Amamra and N. Aouf, "Robust and Sparse RGBD Data Registration of Scene Views," in *2013 17th International Conference on Information Visualisation*, 2013, pp. 488–493.

[38] C. G. M. P. A. P. S. S. Rita Cucchiara, "Improving Shadow Suppression in Moving Object Detection with HSV Color Information."

[39] M. S. Mahmoud, "Resilient linear filtering of uncertain systems," *Automatica*, vol. 40, no. 10. pp. 1797–1802, 2004.

[40] L. Xie, L. Lu, D. Zhang, and H. Zhang, "Improved robust H2 and H∞ filtering for uncertain discrete-time systems," *Automatica*, vol. 40, no. 5. pp. 873–880, 2004.

[41] D. Nguyen-Tuong and J. Peters, "Model learning for robot control: a survey.," *Cogn. Process.*, vol. 12, no. 4, pp. 319–40, Nov. 2011.